\documentclass[twocolumn]{cinc}
\usepackage{graphicx}
\usepackage{comment}
\usepackage{amsfonts}
\usepackage{amsmath,bm}
\usepackage{bbm,color}
\usepackage{multirow}
\usepackage{tabularx}
\usepackage[T1]{fontenc}
\usepackage{aecompl}
\begin{document}
\bibliographystyle{cinc}

% Keep the title short enough to fit on a single line if possible.
% Don't end it with a full stop (period).  Don't use ALL CAPS.
\title{Early Prediction of Sepsis From Clinical Data \\via Heterogeneous Event Aggregation}

% Both authors and affiliations go in the \author{ ... } block.
% List initials and surnames of authors, no full stops (periods),
%  titles, or degrees.
% Don't use ALL CAPS, and don't use ``and'' before the name of the
%  last author.
% Leave an empty line between authors and affiliations.
% List affiliations, city, [state or province,] country only
%  (no street addresses or postcodes).
% If there are multiple affiliations, use superscript numerals to associate
%  each author with his or her affiliations, as in the example below.

\author {Luchen Liu$^{*}$, Haoxian Wu$^{*}$\thanks{* The two authors have equal contribution to this work.}, Zichang Wang, Zequn Liu, Ming Zhang \\
\ \\ % leave an empty line between authors and affiliation
  Department of Computer Science, Peking University, Beijing, China }

\maketitle

% LaTeX inserts the ``Abstract'' heading in the proper style and
% sets the text of the abstract in italics as required.
\begin{abstract}
Sepsis is a life-threatening condition that seriously endangers millions of people over the world. Hopefully, with the widespread availability of electronic health records (EHR), predictive models that can effectively deal with clinical sequential data increase the possibility to predict sepsis and take early preventive treatment.
However, the early prediction is challenging because patients' sequential data in EHR contains temporal interactions of multiple clinical events. And capturing temporal interactions in the long event sequence  is hard for traditional LSTM. 
Rather than directly applying the LSTM model to the event sequences, our proposed model firstly  aggregates heterogeneous clinical events in a short period and then captures temporal interactions of the aggregated representations with LSTM. 
Our proposed Heterogeneous Event Aggregation can not only shorten the length of clinical event sequence but also help to  retain temporal interactions of both categorical and numerical  features of clinical events in the multiple heads of the aggregation representations.
%In this model, we replace the input-embedding with an interpolation of the sparse raw data and follow the subsequent structure (positional encoding, self-attention mechanism, and dese interpolation) of SAnD (Simply Attend and Diagnose) to get low-dimensional representation vectors of patients. 
%In addition, we adopt oversampling technics, such as SMOTE, in the patient representation space to avoid a lack of positive samples for training models. 
%We randomly divided the training data into train, test, validation sets with the ratio size of 7:3:1. The test set serves as the evaluation of our model and we use a validation set to get the estimated utility score.
In the PhysioNet/Computing in Cardiology Challenge 2019 \cite{Reyna2019}, with the team named PKU\_DLIB, our proposed model, in high efficiency, achieved utility score (0.321) in the full test set .
   %25 lines of text. Leave 2 line spaces at the bottom of the abstract 

\end{abstract}
% LaTeX inserts the extra space here automatically.

\section{Introduction}
Sepsis is a life-threatening condition that arises when the body's response to infection causes injury to its tissues and organs. And the early prediction of the   sepsis onset is important for physicians to take early preventive treatment. However, sepsis prediction is a  difficult task, because there are complex sepsis risk factors  including the age of patient, immune system weakness,  complication (e.g. cancer, diabetes), conditions (e.g. trauma, burns) and so on. 
Hopefully, with the help of the widespread availability of electronic health records (EHR), well-designed predictive models, which can effectively make use of clinical sequential data, will be able to increase the sepsis prediction performance.

%data difficulties 
The early sepsis prediction is challenging because patients' sequential data in EHR contains temporal interactions of multiple clinical events \cite{Mio2017deep,qian2017topic}. The interactions of multiple clinical events include event co-occurrence in a short period (e.g. two related symptoms occur together) and event temporal dependency at large time-scale (e.g. A vital signal abnormally arises several hours after certain drug injection). One possible solution is directly applying deep sequential models, such as LSTM \cite{hochreiter1997long}, Transformer\cite{vaswani2017attention}, on the clinical event sequence. However, capturing temporal interactions in the long event sequence is hard for traditional LSTM because the length of clinical sequences exceeds the modeling ability of LSTM.

%traditional methods
Rather than directly applying the LSTM model to the event sequences, some works design hierarchical neural networks to model the long sequence\cite{che2018hierarchical}. For example, aggregating events in a short period into a vector helps to shorten the original long sequence\cite{liu2019learning}. However, the information of each kind of clinical events is mixed in the aggregation vector, so temporal
interactions of these events are hard to capture.

%our idea
To address these issues, our proposed model firstly  aggregates heterogeneous clinical events in a short period and then captures temporal interactions of the aggregated representations with LSTM. 
The Heterogeneous Event Aggregation module can not only shorten the length of clinical event sequence but also help to  retain temporal interactions of both categorical and numerical  features of clinical events in the multiple heads of the aggregation representations. The separated clinical information in different heads makes it easier to capture event temporal interactions in different aggregation vectors.
Experiments on the PhysioNet/Computing in Cardiology Challenge 2019 show that our proposed model is effective and efficient compared to traditional methods.
The contributions of this work is summarised as following:
\begin{itemize}
\item We propose a model to capture temporal interactions among multiple types of clinical event streams from EHR data for early sepsis prediction.
\item The proposed heterogeneous event aggregation module can reduce the length of long clinical event sequences and retain their temporal interactions.
\item Our proposed model achieves good prediction performance and time efficiency.
\end{itemize}

\section{Dataset and Preprocessing}

\subsection{Dataset}
The EHR data provided publicly for this challenge is sourced from two separate ICU, containing 20000 and 20643 records respectively. Each record is made up of hourly clinical data for a specific patient. Each row represents a single hour's data with 40 variables and an additional label indicating whether the patient will get sepsis within 6 hours. With a positive sample proportion of 7.21\%, there are 2932 sepsis patients in total. Sepsis and normal patients are divided into train and test set at the same ratio respectively. 5-fold cross validation is established over the train set.

\subsection{Data Preprocessing}
In this competition, our goal is to make an early sepsis detection within 6 hours for every time-point without causal model. For a patient's record, We use a fix-length sliding window, with 1 hour step, to sample fix-length records $x_t = r_{t-L+1 : t}$ (zero filling if $t < L$).
\par The 40 columns of the record contain 37 numerical variables and 3 binary variables. At the preprocessing stage, we relabel the 3 variables of two-categorical from 0 to 6 (6 clinical categories and one empty category for NaN). Finally we reorder the categorical variables to the last columns. As for numerical variables, in order to make the deep model converge easily, for each numerical variable, z-score normalization is applied. 

\section{Proposed Model}
In this section, Heterogeneous Event Aggregation (HEA) module is designed to effectively capture the interaction information among the heterogeneous clinical events. The motivations of HEA are listed following: (1) Modeling interaction of both categorical and numerical heterogeneous clinical events from their embedding.
(2) Grouping events into multiple heads in different aspects.
(3) Shortening the length of clinical event sequence.
\par After the feature is extracted from HEA, the temporary dependency is captured by bidirectional LSTM. At last, the final outputs of 2 direction is sum up and forwarded to a single dense layer with sigmoid activation to get the detection.

\begin{figure}[h]
\includegraphics[width=8cm]{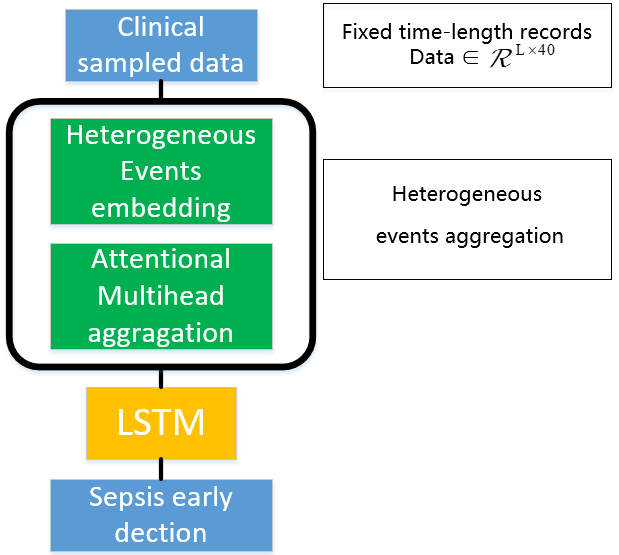}
\caption{The architecture of our proposed model}
\label{FIGURE1}
\end{figure}
The proposed architecture is shown in Figure 1. Given a sequential clinical record $(X_1,X_2,X_3...X_L)$ $X_i \in {\mathbb{R}}^{40}$, our objective is to generate an early prediction of sepsis for the last time step $X_L$. Heterogeneous Event Aggregation (HEA) module is composed of two parts, Heterogeneous Events Embedding and Attentional Multihead Aggregation, which are specifically explained as follows.
\subsection{Heterogeneous Events Embedding} Given the sequential clinical data, the first step of our model is to generate the embedding that can be used to capture the interaction representation among the heterogeneous clinical events \cite{liu2018learning}. For each time step $X_t \in {\mathbb{R}}^{40}$ (the first 37 columns are numerical variables, and the last 3 are categorical variables). Randomly initialized numerical event vector book $W_{ne} \in {\mathbb{R}}^{37\times d}$, categorical event lookup table $W_{ce} \in {\mathbb{R}}^{7\times d}$ and value vector table $W_vn \in {\mathbb{R}}^{37 \times d}$ are generated. The embedding for $X_t$ is then generated as:

\begin{align}
\setlength{\abovedisplayskip}{3pt}
E_t & = Mask(Concat(En_t, Ec_t)) \\
En_t & = W_{ne} + X_t[:37]{W_vn} \\
%numerical embedding
Ec_t & = Embedding\_lookup(X_t[37:], W_{ce}) \\
%categorical embedding
K_t & = Concat(W_{ne}, Ec_t)
%event key
\setlength{\belowdisplayskip}{3pt}
\end{align}
Mask function is used to mask the event embedding to zero if the corresponding variable is default. d is the dimensional numbers of embedding. $E_t \in {\mathbb{R}}^{40 \times d}$ is combination of both numerical and categorical embedding, $En_t \in {\mathbb{R}}^{37\times d}$ is the numerical embedding, $Ec_t \in {\mathbb{R}}^{3\times d}$ is the categorical embedding, $K_t \in {\mathbb{R}}^{40\times d}$ is the Key matrice of both numerical and categorical events. 

\subsection{Attentional Events Aggregation}
\begin{figure}[h]
\includegraphics[width=8cm]{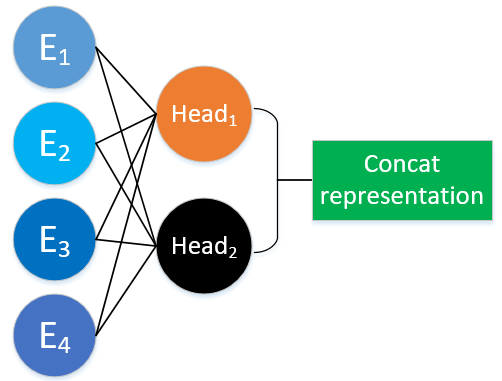}
\caption{An example of two-heads events aggregation}
\label{FIGURE2}
\end{figure}

\par It is difficult to extract the information through the long sequential data for the two reasons:
(1) Within the long sequential record, the interaction among heterogeneous events could be complex, it is difficult to capture the dynamic interaction events representation;
(2) The total dimensional numbers of the heterogeneous events embedding could be disastrously vast, making it impossible for the sequential model to capture the temporal representation.
\par To effectively capture the dynamic heterogeneous events representation, we propose attentional multihead aggregation. Given a time step events embedding $E_t \in {\mathbb{R}}^{40\times d}$, M randomly initialized mask-vectors are generated. Different masks are used to capture different aspects of events-interaction at $time_t$ with attention-based mechanism. At last, all heads are concatenated to produce the eventual aggregation representation. The details of the aggregation are shown as follows:
\begin{align}
\textbf{h}_i = \sum\limits_{j=1}^{40}&attscore_i(K_t[j],\textbf{m}_i)(W_{vi}E_t[j])\\
attscore_i(\textbf{k},\textbf{m}) & = Softmax((W_{ki}\textbf{k}) \cdot \textbf{m}) \\
\textbf{a} &= Concat(\textbf{h}_1, \textbf{h}_2, ..., \textbf{h}_m)
\end{align}

Where $\textbf{h}_i \in {\mathbb{R}}^{d}$ is the $i^{th}$ head of aggregation representation, $\textbf{a} \in {\mathbb{R}}^{M\cdot d}$ concatenates all heads, $attscore_i$ is the attention mechanism of the $i^{th}$ Head, getting the aggregation proportion of each event with calculating the dot product between events and Mask. Each head could capture its own concerned events information with its corresponding formed transform matrices $W_k$, $W_v$ and Mask vector \textbf{m}. An example of two-heads aggregation is shown as Figure 2.

\subsection{Sequential Model and Prediction} For each time step, we get an aggregation representation. Given $X \in {\mathbb{R}}^{L\times40}$, a temporal events aggregation representation $\textbf{a} \in {\mathbb{R}}^{L\times (M \cdot d)}$ is captured. As LSTM is successfully used in sequential data, we pass on A to one-layer bidirectional LSTM module. We sum up the last-time outputs of both forward and backward units and get the logits through a single dense layer with sigmoid activation. Our objective is a binary classification, we use cross entropy:
\begin{equation}
    L(y, \hat{y}) = -(\hat{y}\cdot\log(y) + (1-\hat{y})\cdot log(1-y))
\end{equation}

\section{Experiment Results}
\begin{table}[htbp]
\caption{local partitioned test metrics for different model}
\vspace{4 mm}
\small
\centerline{\begin{tabular}{cccc} 
\hline
\hline
Model    & AUC &APC &Utility score\\ 
\hline
MLP   &0.7723   &0.0711  &0.2413  \\
Transformer   &0.8013    &0.0923   & 0.3619       \\
LSTM    &0.8165     &0.1040     &0.3723     \\
1-head-HEA-Transformer  &0.8092 &0.0955 &0.3775\\
1-head-HEA-LSTM     &0.8241     &0.1120     &0.3877     \\
8-heads-HEA-Transformer    &0.8264    &0.1236    &0.3968   \\
8-heads-HEA-LSTM   &0.8400   &0.1307   &0.4096   \\
16-heads-HEA-LSTM  &\textbf{0.8410}    &\textbf{0.1314}    &\textbf{0.4126}   \\
\hline\hline
\end{tabular}}
\end{table}

\begin{table*}[htbp]
\caption{local cross-validation metrics for HEA}
\vspace{4 mm}
\centerline{\begin{tabular}{cccc} 
\hline
\hline
Model    & AUC &APC & Utility score(average/major vote/any vote)\\ 
\hline
8-heads-HEA-Transformer   &0.8186   &0.1049  &0.3641/0.3635/0.3658  \\
8-heads-HEA-LSTM   &0.8206    &0.1031   & 0.3656/0.3613/0.3643       \\
16-heads-HEA-LSTM    &\textbf{0.8224}    &\textbf{0.1052}    &0.3817/0.3750/\textbf{0.3830}   \\
\hline\hline
\end{tabular}}
\end{table*}

\subsection{Implementation Details}
The provided data is divided into train and test sets at the ratio of 7 to 3. To conduct a further experiment, we divide the data into train and held sets at the ratio of 9 to 1, and then 5-fold cross validation is established. To evaluate our proposed module, we directly use a MLP prediction model, Transformer and LSTM with dense layer embedding, set various numbers of heads to gain different models. The results show that our proposed model, with high efficiency, can obviously improve the performance. We keep L as 24 (one day) in all experiments.\\
\textbf{Multihead aggregation}: We keep dimensional numbers(d) as 16 and set heads to be 1, 8 and 16 as 3 different aggregation modules.
\par We measure Area Under the Receiver Operating Characteristic curve (AUC) and Area Under The Precision Recall Curve (APC) as our evaluation metrics. What's more, the utility score function defined in CinC2019 challenge is used as the extra metric.

\subsection{Baseline}
\textbf{MLP}: Without considering the temporal information, Multi-layer perception can be used to directly model the raw record.

\textbf{Dense Layer Embedding}: Given a one-time sampled record \textbf{x}, the dense representation layer use a single dense layer with activation to generate the representation $\hat{\textbf{x}} \in \mathbb{R}^{d}$.
%\begin{equation}
%   \hat{\textbf{x}} = Wx + b
%\end{equation}

\textbf{Transformer /\ LSTM}: Conventional sequential model Transformer and LSTM use a single dense layer embedding to capture the events representation. 

\subsection{Result}
We firstly conduct an experiment for different models over train and test sets. The results over the test set are shown as Table 1.
It should be noticed that all the metrics on both Table 1 and Table 2 are based on the locally partitioned test set from the public dataset.

The result shows that heterogeneous events aggregation modules could improve the metrics obviously, then we have the further experiment over 5-fold cross validation. The results over the held set are as Table 2 shows.

What's more, our proposed model is in high efficiency, for it just needs to calculate the attention score between events and multiple heads. With a GeForce GTX 1080 10G, our proposed model with 16 heads cost 10 minutes per epoch to train over 10240000 training samples.

\par In thePhysioNet/Computing in Cardiology Challenge 2019, we got the results of score (0.402, 0.386, -0.169) on the test set A,B and C, with the overall utility score of 0.321, ranking the 13th out of 78 teams.

\section{Conclusion}
We proposed an attention-based sequential representation model to do early sepsis prediction from clinical data.
Our proposed model includes two main parts: clinical events interaction extraction with heterogeneous events aggregation and temporal interaction capture with LSTM. Experiments in the PhysioNet/Computing in Cardiology Challenge 2019 show that the heterogeneous event aggregation module can  shorten the length of clinical event sequence for better temporal dependency modeling, and the separated storage strategy of aggregation representation with different heads retains temporal interactions of events.

\section*{Acknowledgments}  
This paper is partially supported by
National Key Research and Development Program of China with Grant No. 2018AAA0101900,
Beijing Municipal Commission of Science and Technology under Grant No.
Z181100008918005, 
and the National Natural Science Foundation of China (NSFC Grant No. 61772039 and No. 91646202).
% This section is not numbered.
% 

% LateX generates the ``References'' heading automatically and switches
% to 9 point type for the bibliography.  Please  use BibTeX and
% follow the examples in the sample 'refs.bib' file to enter your references.
\bibliography{refs}

% If you don't use BibTeX (why not?) , comment out or remove the previous
% line, and uncomment the following lines up to the ``}\end{bibliography}''
% line below:
%\begin{thebibliography}{99}{ %\small
% \bibitem{tag} (General form) J. K. Author, ``Name of paper,''
%   \emph{Abbrev. Title of
%   Periodical}, vol. x, no. x, pp. xxx--xxx, Abbrev. Month, year. 

% \bibitem{ito}  M. Ito et al., ``Application of amorphous oxide TFT to
%   electrophoretic display,'' \emph{J. Non-Cryst. Solids}, vol. 354, no. 19,
%   pp. 2777--2782, Feb. 2008.
  
% \bibitem{fardel}  R. Fardel, M. Nagel, F. Nuesch, T. Lippert, and
%   A. Wokaun, ``Fabrication of organic light emitting diode pixels by
%   laser-assisted forward transfer,'' \emph{Appl. Phys. Lett.}, vol. 91,
%   no. 6, Aug. 2007, Art. no. 061103.
  
% \bibitem{buncombe} J. U. Buncombe, ``Infrared navigation Part I: Theory,''
%     \emph{IEEE Trans. Aerosp. Electron. Syst.}, vol. AES-4, no. 3,
%     pp. 352--377, Sep. 1944.
      
% Uncomment the following line if you are not using BibTeX.
%}\end{thebibliography}

% LaTeX inserts the ``Address for correspondence'' heading.
\begin{correspondence}
Ming Zhang\\
1628, No.1 Science Building, Peking University, Beijing, China\\
mzhang\_cs@pku.edu.cn
\end{correspondence}

\end{document}